\documentclass{article}
\usepackage{booktabs}
\usepackage[final]{corl_2026}

\usepackage[shortlabels]{enumitem}

\usepackage[normalem]{ulem}

\usepackage{amssymb}

\usepackage{amsthm}

\usepackage[makeroom]{cancel}

\usepackage{amsmath} 
\newtheorem{theorem}{Theorem}[section]

\newtheorem{proposition}[theorem]{Proposition}
\newtheorem{remark}[theorem]{Remark}

\usepackage{amsfonts}
\newcommand{\R}{\mathbb{R}}

\newcommand{\mb}[1]{\mathbf{#1}}
\newcommand{\bs}[1]{\boldsymbol{#1}}

\DeclareMathOperator*{\argmin}{argmin}

\usepackage{tikz}

\usepackage{fancyhdr}
\usepackage[most]{tcolorbox}

\newtcolorbox{coreinsight}{
    colback=gray!5,     
    colframe=black,     
    width=\textwidth,   
    arc=2mm,            
    auto outer arc,
    boxrule=0.5pt,      
    left=15pt,          
    right=15pt,
    top=10pt,
    bottom=10pt,
    fontupper=\itshape, 
    borderline west={3pt}{0pt}{black} 
}

\usepackage{array}
\usepackage[table,x11names,dvipsnames,svgnames]{xcolor} 

\usepackage{textcomp} 

\definecolor{myLinkBlue}{RGB}{0,0,160}
\definecolor{myCiteGreen}{RGB}{0,110,0}
\definecolor{myUrlBlue}{RGB}{0,80,180}

\hypersetup{
    colorlinks=true,
    linkcolor=myLinkBlue,
    citecolor=myCiteGreen,
    urlcolor=myUrlBlue
}

\title{
\Large
MARCH: \uline{M}odel-\uline{A}ssisted \uline{R}einforcement Learning for the Perceptive \uline{C}ontrol of \uline{H}umanoids over Sparse Footholds
\vspace{-1em}
}

\author{
  Codrin Crismariu and Ryan K. Cosner\\
  Department of Mechanical Engineering\\
  Tufts University,  
  Medford, MA, United States\\
  \texttt{codrin.crismariu@tufts.edu, ryan.cosner@tufts.edu}
}

\begin{document}
\maketitle

\vspace{-3em}

\begin{figure}[htbp] 
    \centering
    \includegraphics[width=0.95\linewidth]{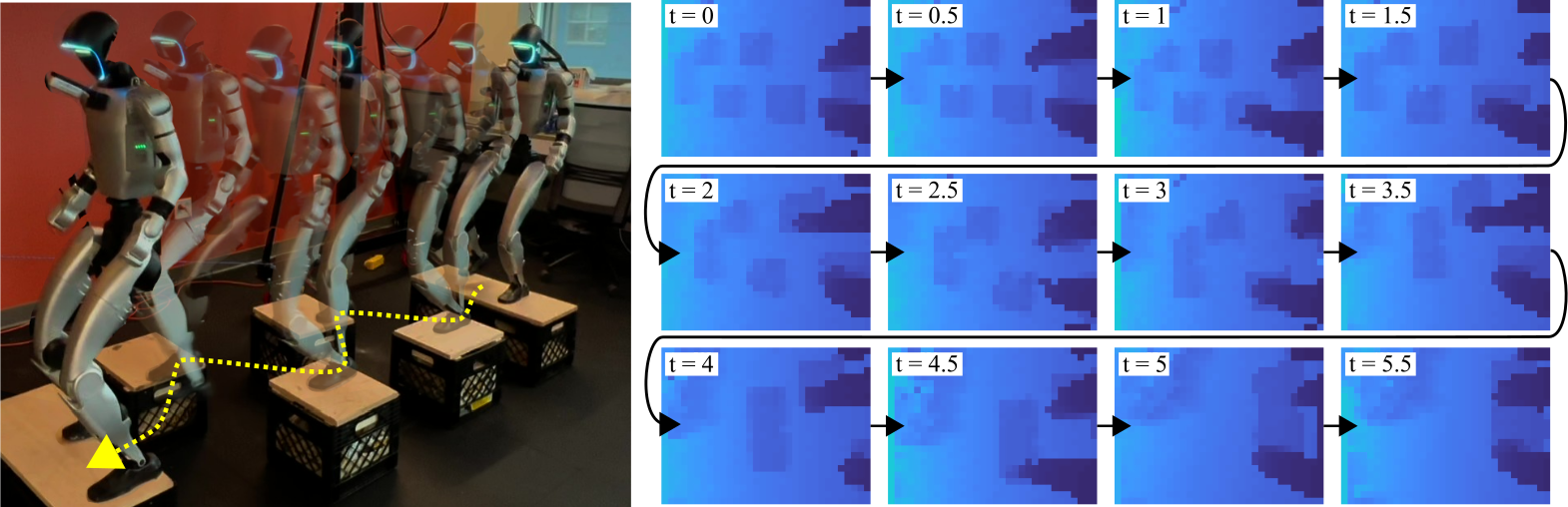}
    \caption{Unitree G1 humanoid traversing sparse footholds with a vision-based, model-assisted reinforcement learning (RL) control policy. \textbf{(Left)} Physical experiment snapshots; \textbf{(Right)} depth images for control-policy feedback. Experiment video: \href{https://www.youtube.com/watch?v=pXjy-zPCRPA}{youtube.com/watch?v=pXjy-zPCRPA}.}
    \label{fig:hardware}
    \vspace{-1em}
\end{figure}


\begin{abstract}
Perceptive bipedal locomotion over sparse terrain remains a difficult challenge: model-based methods are precise but brittle to uncertainty, while model-free methods are robust but struggle to discover the precise, constrained motions required for precise locomotion where small errors can cause catastrophic failures. 
We propose a model-assisted reinforcement learning (RL) framework that combines both perspectives in three steps: 
(1) generate a reference trajectory using simplified models; (2) train a privileged teacher policy with Lyapunov-guided heuristic reward shaping around that reference; and (3) distill the teacher into a vision-based student policy.  
We show that this model-assistance procedure produces physically grounded locomotion, improving sample efficiency, reducing the need for a complex learning curriculum, and achieving smoother locomotion behavior alongside stepping stone performance comparable to a model-free baseline. We validate our approach in simulation and demonstrate a deployment on a Unitree G1 humanoid robot navigating sparse footholds with lateral constraints.
\end{abstract}

\keywords{Reinforcement learning for physical robot control,  Model-based and model-free learning for robotic control, Bipedal locomotion} 


\section{Introduction} 
\label{sec:intro}
The promise of bipedal robotics is to \textit{enable robots to go wherever humans can go}. However, while current methods have successfully created robust  locomotion policies that enable walking on nominal terrain, they struggle to generalize to scenarios like sparse footholds with lateral constraints. Developing policies to navigate these environments currently demands complex training curricula and sampling-intensive training.

Existing methods for achieving safe locomotion over sparse terrain range from model-based to model-free. Traditional model-based methods typically rely on \textit{a priori} known system dynamics models and known environment maps \cite{nguyen20163d,csomay2021episodic,grandia2021multi}. The assumptions of a known dynamics model and safe foot placement map allow them to generate rigorous safety guarantees, but these assumptions struggle to generalize to real-world deployment and new scenarios, especially when the robot must rely on high dimensional visual measurements to identify sparse footholds. Model-free approaches typically involve either (1) developing a map from sensor feedback and learning a locomotion policy that enables safe traversal over that map \cite{miki2022learning, ben2025gallant} or (2) directly learning a locomotion policy that incorporates visual information as an input \cite{zhang2026rpl}. Recent work explored a middle ground between model-based and data-driven methods by using models to guide reinforcement learning (RL) in the form of control Lyapunov function RL (CLF-RL) where an approximate model is used in a CLF reward function \cite{li2026clfrl}. CLF-RL shows promise in improving performance and RL sample efficiency, with demonstrations on safe locomotion involving forward foothold constraints \cite{dai2026walk}, but has yet to be demonstrated in scenarios with lateral foothold constraints; this work provides that extension.

To achieve vision-based locomotion over sparse footholds with lateral constraints, we extend \cite{dai2026walk} by incorporating a receding-horizon safe footstep planner and adding a transformer and mixture density network (MDN) \cite{astonpr373} to the locomotion policy in order to plan multiple steps, increase the utility of the proprioception measurements, and allow the policy to select between modes representing different foot placements. An overview of this framework can be seen in Figure \ref{fig:overview}.

The core contributions of this work are: (1) a model-assisted RL framework that extends \cite{dai2026walk} to incorporate lateral foot placement requirements by introducing a receding horizon planner, a transformer, and an MDN output, (2) simulation comparisons to model-free methods that demonstrate our method's improved sample efficiency and smooth locomotion with an ablation study showing the utility of our policy architecture, and (3) an experimental demonstration of our framework on a Unitree G1 humanoid robot walking over terrain with variable gaps and stepping stone placement, demonstrating locomotion over sparse terrain with lateral constraints on hardware.

\section{Related Work}

Model-based methods have tackled the problem of locomotion over sparse footholds and provided rigorous guarantees of safety using hybrid system models. Notably, \cite{nguyen2015safety} presented a control barrier function (CBF \cite{ames2019control})-based safe foot placement method for a planar biped which was later extended to a three dimensional biped in \cite{nguyen20163d}, with both works limited to simulation. Later hardware deployment of these methods \cite{csomay2021episodic} required data-driven model refinements, removing their theoretical guarantees. Similar model-based methods for quadrupedal robots leverage state-constrained model predictive control (MPC) to achieve safe foot placement when the environment height map is known \cite{grandia2021multi} and when it is constructed from perception data \cite{grandia2023perceptive}. Generally, model-based methods for safe locomotion over sparse footholds enable rigorous theoretical guarantees, but rely on conservative modeling and perception assumptions that are difficult to satisfy on hardware.

Meanwhile, model-free, learning-based  methods have enabled robust locomotion over nominal terrains \cite{zhuang2024humanoidparkourlearning}. While prior work shows the efficacy of RL for quadrupedal locomotion over a diverse set of terrains \cite{cheng2023extremeparkourleggedrobots,agarwal2023legged, lee2020learning}, 
similar deployment for humanoids presents a more difficult problem due to the increased instability of bipeds.
Recent solutions for humanoids either avoid perception and rely on commanded foot placement \cite{duan2022sim}, incorporate perception information by constructing maps \cite{sun2025dpldepthonlyperceptivehumanoid}, or operate directly on perception measurements \cite{sun2025learningperceptivehumanoidlocomotion, zhuang2024humanoidparkourlearning, zhu2026hikingwildscalableperceptive}. The complexity of these learning problems generally limit their results to obstacles that can be overcome with general flat-terrain robustness or obstacles that do not impose a lateral constraint (e.g., stairs) and have prevented these methods from being deployed on stepping stones that are placed far apart where a poorly-placed step might result in catastrophic behavior. Only recent results have been able to achieve this behavior, but require large amounts of compute and highly structured curricula to find a viable solution in the poorly-conditioned reward landscape \cite{zhang2026rpl}. 


To improve the computational efficiency of this RL problem, \cite{li2026clfrl} proposed the CLF-RL method that incorporates a model-guided reward function based on control Lyapunov functions (CLFs) \cite{artstein1983stabilization} into a traditional RL framework. This method uses model-based reference trajectories to guide policy learning by rewarding policies that produce motions similar to a safe model-based reference trajectory. CLF-RL alongside teacher-student distillation \cite{lee2020learning} has been shown to improve computational efficiency in learning complex behaviors like running \cite{olkin2026chasingautonomydynamicretargeting} and forward-constrained locomotion over sparse footholds in \cite{dai2026walk}. Similarly, \cite{janwani2026navigaitnavigatingdynamicallyfeasible} incorporates a model-based gait library which is used to guide RL to develop more general locomotion policies. Despite the success of these model-assisted results, they have yet to be applied to a stepping stone problem involving lateral constraints.
This paper uses Lyapunov-guided heuristic reward shaping to learn walking on sparse footholds that induce both forward and lateral constraints. 


\vspace{-0.5em}
\section{Preliminaries} \label{sec:preliminaries}

\vspace{-0.5em}

Next, we introduce our various system models and student-teacher reinforcement learning paradigm. 

\subsection{System Models}

\textbf{General Walking as a Hybrid System:} We model general bipedal walking as a hybrid system that incorporates both continuous motion and the discrete changes resulting from the impacts. We denote this system as $\Sigma = (\mathcal{D}, \mathcal{U}, \mathcal{G}, \Delta, \mb{f}, \mb{g})$ where $\mathcal{D}$ is the set of continuous domains, $\mathcal{U}\subset \R^{n_u}$ is the control input space, $\mathcal{G}\subset\mathcal{D}$ represents the guards corresponding to impact events, $\Delta: \mathcal{G} \to \mathcal{D}$ defines the discrete reset map, and $\mb{f}$ and $\mb{g}$ define the continuous dynamics of the system. We denote the state as $\mb{x} \triangleq [ \mb{q}^\top, \; \dot{\mb{q}}^\top]^\top \in \mathcal{TQ}$ where $\mb{q}$ is the generalized system coordinate and the system input as $\mb{u} \in \mathcal{U}$. In general, we assume that we do not have access to these precise dynamics:
\begin{align} \label{eq:hybrid_sys}
    \begin{cases}
        \dot{\mb{x}} = \mb{f}(\mb{x}) + \mb{g}(\mb{x}) \mb{u}, & \mb{x}\in \mathcal{D} \setminus \mathcal{G},\\
        \mb{x}^+ = \Delta(\mb{x}^-), & \mb{x}^- \in \mathcal{G}. 
    \end{cases}
\end{align}


\textbf{Discrete-Time Single Integrator:} While the hybrid system \eqref{eq:hybrid_sys} describes the general motion of the bipedal robot, its complexity makes it difficult to use for planning foot placement over sparse terrain. Alternatively, we consider a simplified discrete-time single integrator model of the stance foot location, $\mb{p}_\textup{f} \in \R^2$: $\mb{p}_{\textup{f}, k+1} = \mb{p}_{\textup{f},k} + \bs{\ell}_k$, where $\bs{\ell}_k$ is the change in stance foot location that occurs between steps. This dramatically simplified model cannot be used directly for control and overlooks important system complexity, but its simplicity enables rapid planning and constraint checking to ensure that the planned stance foot locations are safe ($\mb{p}_{\textup{f}, k} \in \mathcal{L}_\textup{step feas.}$) and that the planned step length is feasible ($\bs{\ell}_k \in \mathcal{L}_{\textup{step bound}}$). 

\textbf{Hybrid Linear Inverted Pendulum Model \cite{xiong_hlip_2022}:} To capture the hybrid, underactuated nature of walking and characterize the  swing phase motion, we model the bipedal robot as a hybrid linear inverted pendulum (HLIP) with discrete transitions at each impact \cite{xiong_hlip_2022}. This model simplifies the hybrid dynamics in \eqref{eq:hybrid_sys} by assuming that the system has point-feet, massless legs, no ankle torque, a constant center of mass (CoM) $z$-height, and all mass concentrated at the CoM, resulting in an underactuated system 
where the horizontal CoM position relative to the stance foot, $[r_x, \; r_y]$, cannot be directly regulated and must instead be controlled via the step-to-step dynamics. This HLIP model is linear and characterizes both the discrete step-to-step motion and continuous flow of the CoM between steps which can be decoupled into two orthogonal planes (i.e., $x$ and $y$).  



To describe the HLIP model, let $\boldsymbol{\xi}_x = [r_x, \; v_x]^\top$ be the  state vector where $v_x\in \R$ is $x$-direction CoM velocity. The continuous dynamics of the HLIP model are: 
\begin{align}
    \dot{\boldsymbol{\xi}}_x = \frac{d}{dt}\begin{bmatrix} r_x\\
    v_x
    \end{bmatrix} = \underbrace{\begin{bmatrix}
        0 & 1\\
         \frac{g}{z_0} & 0 \end{bmatrix}}_{A}\bs{\xi}_x, && \textup{ yielding the intra-step dynamics: }  \quad \quad \bs{\xi}_x(\tau) = e^{A\tau}\bs{\xi}_x^+, \label{eq:lip}\\
         &&\textup{ and the post- to pre-impact dynamics: } \quad \quad \bs{\xi}_x^- = e^{AT}\bs{\xi}_{x}^+, \label{eq:hlip-step-to-step}
\end{align}
where $g$ is gravitational acceleration, $z_0$ is the $z$-height of the CoM, $\tau$ is the time within the current swing phase, and $T$ is the step period. 
The reset map accounts for the step length $\ell_x$, resulting in: 
\begin{align}
    \bs{\xi}_x^+ = \bs{\xi}_{x}^- + \underbrace{\begin{bmatrix}
        0 \\ - 1
    \end{bmatrix}}_{B} {\ell}_x, && \textup{which yields the step-to-step dynamics: } \bs{\xi}_{x,k+1}^{-} = e^{AT} \bs{\xi}_{x,k}^-  + e^{AT}B \ell_{x,k}.   \nonumber 
\end{align}
We then identically define $\bs{\xi}_y$ for the CoM in the $y$-direction and use this model to plan, predict, and control CoM motion in the $(x, y)$ plane.

\subsection{Reinforcement Learning and Teacher-Student Distillation}
We formulate the task of sparse foothold traversal as a partially observable Markov decision process (POMDP), defined by the tuple $\mathcal{M} = (\mathcal{S}, \mathcal{A}, \mathcal{O}, \mathcal{T}, \mathcal{Z}, \mathcal{R}, \gamma)$ where $\mathcal{S}$ is the set of states, $\mathcal{A}$ is the set of actions, $\mathcal{O}$ is the set of observations, $\mathcal{T}(\mb{s}'|\mb{s}, \mb{a}) $ is the state transition probability of moving from state $\mb{s}$ to state $\mb{s}'$ by taking action $\mb{a}$, $\mathcal{Z}(\mb{o}|\mb{s})$ is the probability of observing $\mb{o}$ given that the system is at state $\mb{s}$, $\mathcal{R}(\mb{s}, \mb{a})$ is the instantaneous reward function, and $\gamma \in [0, 1]$ is the discount factor. The agent seeks to maximize the expected discounted cumulative reward: 
\begin{align}
    J(\bs{\pi}_{T,\theta}) = \mathbb{E}_{ \bs{\pi}_{T,\theta}} \left[ \sum_{k = 0 }^K \gamma^k \mathcal{R}(\mb{s}_k, \mb{a}_k) \right] \label{eq:reward}
\end{align}
where $\bs{\pi}_{T, \theta}(\mb{a}_k|\mb{o}_k, \mb{o}_k^\textup{priv.})$ is the stochastic policy parameterized by $\theta$ and the expectation is taken over trajectories generated by this policy. 
We optimize the parameters $\theta$ to maximize $J(\bs{\pi}_{T, \theta})$ using proximal policy optimization (PPO) \cite{schulman2017proximal}. 

To improve learning under partial observability, we leverage a teacher-student framework. The teacher policy $\bs{\pi}_{T,\theta}(\mb{a}_k|\mb{o}_k, \mb{o}^{\textup{priv.}}_k)$ is given access to privileged observation information $\mb{o}_k^{\textup{priv.}}$ while the student policy $\bs{\pi}_{S, \phi}(\mb{a}_k|\mb{o}_k)$ has limited sensing information. Once the teacher policy is trained, the student policy is trained to match the teacher under the supervised distillation loss: 
\begin{align}
    \mathcal{L}_{\textup{distill}} = \mathbb{E}_{\bs{\pi}_{T, \theta}} \left[ - \log \left( p_{\bs{\pi}_{S, \phi}}(\mb{a}_{T,k} |\mb{o}_k)  \right) \right]
    \label{eq:distillation_loss}
\end{align}
where $\mb{a}_{T, k}\sim \bs{\pi}_{T, \theta}(\cdot | \mb{o}_k, \mb{o}_k^\textup{priv.})$ is the action sampled from the teacher distribution. 


\section{Methodology} \label{sec:methods}
\begin{figure}[t]
    \centering
    \includegraphics[width=0.99\linewidth]{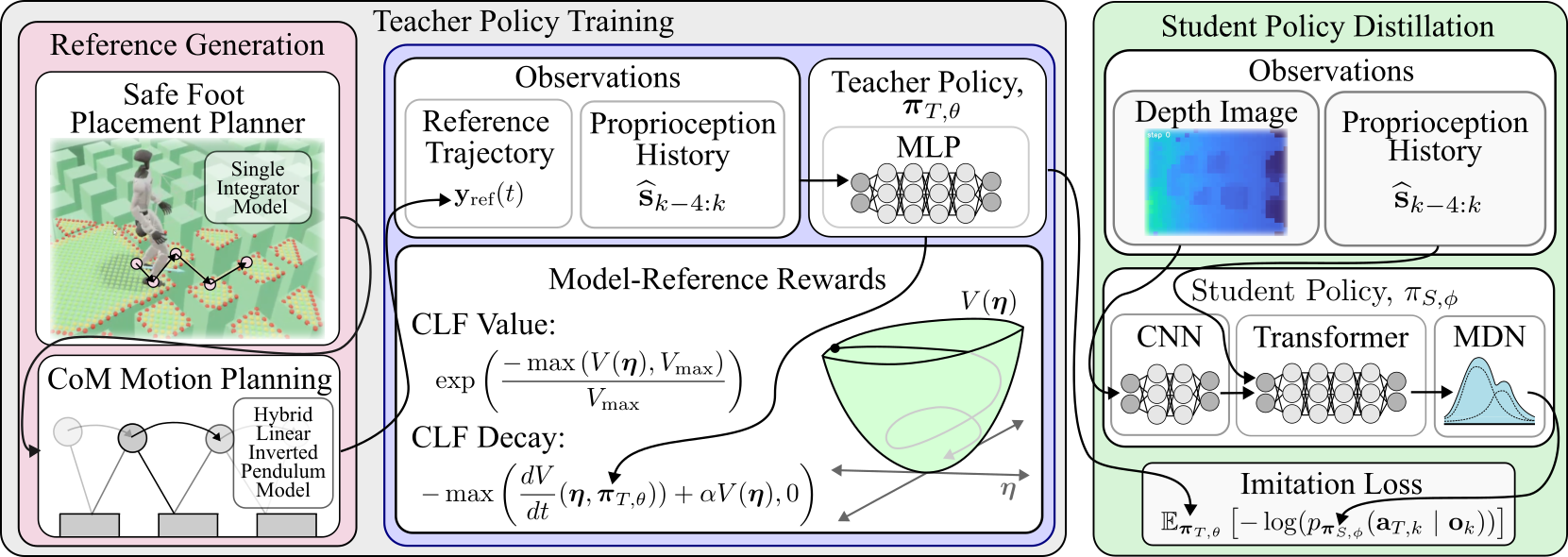}
    \caption{\textbf{Framework Overview}. Our framework involves three core components: \textcolor{red}{\textbf{(red)}} generation of model-based safe reference trajectories, \textcolor{blue}{\textbf{(blue)}} training of a privileged teacher control policy that uses control Lyapunov function (CLF)-inspired rewards in a CLF-RL \cite{li2026clfrl} framework, and \textcolor{olive}{\textbf{(green)}} distillation of the teacher policy into a student policy that only relies on the visual and proprioceptive data accessible during hardware deployment. }
    \label{fig:overview}
    \vspace{-1em}
\end{figure}

Building on the models and student-teacher RL framework in Section \ref{sec:preliminaries},  
our approach develops a control policy via a three-step process which can be seen in Figure \ref{fig:overview}. First, we develop a model-based reference trajectory to identify a region of local optimality for teacher policy. Next, we leverage privileged information in simulation to train a teacher policy using Lyapunov-based reward shaping as in \cite{li2026clfrl,dai2026walk}. Finally, we distill this teacher policy into a student that operates exclusively on visual feedback and proprioception to enable deployment on hardware.




\subsection{Reference Generation}

Here we introduce the methodologies used to generate the safe model-based reference trajectories that are used to guide the  RL reward. Overall, the reference trajectories include the outputs: 
\begin{align}
    \mb{y}_\textup{ref}(t) = (\mb{p}_\textup{f}, \; \mb{v}_\textup{f}, \; \mb{p}_\textup{com}, \; \mb{v}_{\textup{com}}, \; \bs{\varepsilon}_{\textup{pelv}}, \; \dot{\bs{\varepsilon}}_{\textup{pelv}}, \;   \bs{\varepsilon}_\textup{foot}, \;\dot{\bs{\varepsilon}}_{\textup{foot}}, \; \mb{q}_{\textup{upper}}, \; \dot{\mb{q}}_{\textup{upper}}) \label{eq:ref_traj}
\end{align}
where $(\mb{p}_\textup{f}, \; \mb{v}_\textup{f})$ is the swing foot position and velocity,  $(\mb{p}_\textup{com}, \; \mb{v}_{\textup{com}})$ is the CoM position and velocity,  $(\bs{\varepsilon}_{\textup{pelv}}, \; \dot{\bs{\varepsilon}}_{\textup{pelv}})$ are the pelvis orientation and derivatives, $(\bs{\varepsilon}_\textup{foot}, \;\dot{\bs{\varepsilon}}_{\textup{foot}})$ are the swing foot orientation and derivatives, and $(\mb{q}_{\textup{upper}}, \; \dot{\mb{q}}_{\textup{upper}})$ are the joint positions for the robot's upper body. 
The time argument,  $t$, is suppressed for each value on the right-hand side for brevity. 

\textbf{Swing Foot Position Reference, $(\mb{p}_\textup{f}, \; \mb{v}_\textup{f})$: } We first plan safe two-dimensional foot placements and step timing based on a single integrator system in order to safely track a desired velocity, $\mb{v}_D$. Specifically, this plan attempts to solve the finite, discrete-time optimal control problem:
\begin{align}
    \bs{\ell}^*_{k:k+N}, \Delta t_{k:k+N}^* = \argmin_{\substack{\mb{p}_{\textup{f},0:N+1}\in \R^2 \\ \bs{\ell}_{0:N} \in \R^{2},\,\Delta t_{0:N}\in \R_{>0}}} & \sum_{i=1}^{N+1} \Vert \mb{v}_i - \mb{v}_{D} \Vert^2 + c_{\textup{edge}}(\mb{p}_{\textup{f},i}) \label{eq:planner} \\
    \textup{s.t.} \quad & \mb{p}_{\textup{f},i+1} = \mb{p}_{\textup{f},i} + \bs{\ell}_i, \quad \mb{v}_i = \tfrac{\bs{\ell}_i}{\Delta t_{i}}, \quad \forall i \in \{0, \dots, N\}, \nonumber\\
    & \mb{p}_{\textup{f},i} \in \mathcal{L}_{\textup{step feas.}}, \quad \bs{\ell}_{i} \in \mathcal{L}_\textup{step bound}, \nonumber\\
    & \mb{p}_{\textup{f},0} = \mb{p}_{\textup{f},k}. \nonumber
\end{align}
where the cost $c_\textup{edge}(\cdot)$ penalizes the system for stepping near an edge, the first constraint enforces single integrator dynamics, the second constraint determines the average velocity needed to complete the step, the third constraint enforces foot placement in the safe region $\mathcal{L}_{\textup{step feas.}}$, the fourth constraint enforces the step length bound $\mathcal{L}_\textup{step bound}$, and the final constraint assigns the initial condition. 

The edge costs and stepping-stone constraints are highly non-convex making this problem difficult to solve analytically. However, over a short horizon ($N = 4$ steps in our case), a solution can be quickly approximated via random shooting (with 16 greedy rollouts)  over discretized time and safe foot placements. 
Once we have generated a foot-placement and step-timing plan, we construct Bezier curves defining the desired three-dimensional trajectory for the swing foot.

\textbf{CoM Reference, $(\mb{p}_\textup{com}, \; \mb{v}_{\textup{com}})$:} After generating a reference trajectory for the swing foot, we create a reference trajectory for the CoM based on the HLIP model in \eqref{eq:hlip-step-to-step} and the step length and period found by the planner \eqref{eq:planner}.
We interpolate between the pre-impact CoM states using the continuous portion of the HLIP dynamics \eqref{eq:lip}, $\bs{\xi}(\tau)$. 

\textbf{Pelvis Orientation Reference, $(\bs{\varepsilon}_{\textup{pelv}}, \; \dot{\bs{\varepsilon}}_{\textup{pelv}})$:} The yaw reference trajectory includes small heuristic oscillations at the step frequency in roll and pitch, and tracks the commanded yaw rate, $\bs{\omega}_D$. 

\textbf{Swing Foot Orientation Reference, $(\bs{\varepsilon}_\textup{foot}, \;\dot{\bs{\varepsilon}}_{\textup{foot}})$:} The swing foot orientation reference trajectory is chosen to keep the base of the swing foot parallel with the ground to enable flat-foot stepping while aligning the swing foot's yaw with the pelvis.

\textbf{Upper Body Reference, $(\mb{q}_{\textup{upper}}, \; \dot{\mb{q}}_{\textup{upper}})$:} The upper body reference trajectory maintains nominal relative positions and orientations with respect to the CoM. 




\subsection{Reward Shaping and Training the Teacher Policy} \label{subsec:reward_shaping}

Next, we train the teacher policy 
using PPO \cite{schulman2017proximal} with a 
model-guided CLF-RL reward \cite{li2026clfrl}. 

\textbf{Teacher Architecture and Observations: } The teacher's actor and critic are both multi-layer perceptrons (MLPs). The teacher has access to a 5-step history of proprioception data and has privileged access to the reference trajectory \eqref{eq:ref_traj} 
giving it indirect access to ground-truth safe foot placement.

\textbf{Reward Shaping:} 
In simulation, we collect trajectory information to find the error $\bs{\eta}(t) = \mb{y}_\textup{sim}(t) - \mb{y}_\textup{ref}(t) $ which represents the difference between the true values in simulation, $\mb{y}_\textup{sim}(t)$, and the reference values that the system is tracking, $\mb{y}_\textup{ref}(t)$. To incentivize convergence to this reference trajectory, we follow the CLF-RL paradigm  \cite{li2026clfrl} and construct rewards inspired by control Lyapunov functions (CLFs) \cite{artstein1983stabilization}. To do this, we define the Lyapunov-like function $V(\bs{\eta}) \triangleq \frac{1}{2}\bs{\eta}^\top \mb{P} \bs{\eta}$, where $\mb{P}$ is a positive definite matrix that solves the continuous time algebraic Riccati equation 
\cite{astrom2021feedback}, assuming, as in \cite{dai2026walk}, that $\bs{\eta}$ has second-order feedback-linearizable dynamics. 
\begin{remark}
    While it is unlikely that $V$ is a true control Lyapunov function or that the assumption of feedback linearizability holds, this function $V$ can still be used to identify useful local minima in training. Appendix \ref{apdx:clf-rl} provides the conditions for which $V$ is a valid CLF for the single-support swing phase of a fully-actuated bipedal robot. 
\end{remark}

Prior work implemented the CLF constraint, $\frac{dV}{dt}(\bs{\eta}) \leq -c V(\bs{\eta})$, as part of state-feedback controller to ensure that $\bs{\eta}$ converges exponentially \cite{ames2014rapidly}. However, the reliance of these methods on their assumed model introduced fragility to model uncertainty and disturbances \cite{sontag1995characterizations}. To overcome this, the CLF-RL paradigm instead uses $V(\bs{\eta})$ to shape the RL reward function \eqref{eq:reward} by adding two CLF-inspired rewards to the standard locomotion rewards discussed in Sec. \ref{sec:setup}: 
\begin{align} \label{eq:clf_rewards}
    r_\textup{clf} = \exp \left( \frac{-\min \left( V(\bs{\eta}), V_\textup{max}\right) }{ V_\textup{max}}\right) \quad  \textup{ and } \quad r_{\textup{dclf}} = - \max\left( \frac{\frac{dV}{dt}(\bs{\eta}, \mb{a})+ \alpha V(\bs{\eta}) }{V_{\textup{dmax}}}, 0  \right). 
\end{align}
These reward small $V(\bs{\eta})$ values (i.e., accurate reference tracking) and decreasing $V(\bs{\eta})$ values (i.e., convergence to the reference) respectively, with the normalization parameters $V_\textup{max}, V_{\textup{dmax}}>0$. 





%

\subsection{Distillation of the Student Policy}

Finally, we distill a student policy, $\bs{\pi}_{S, \phi}$, by minimizing the negative log likelihood loss \eqref{eq:distillation_loss}.

\textbf{Student Observations:} The student policy receives  a 5-step history of the proprioception data and the desired velocity command as well as the current image from a body-mounted depth camera. The student does not have access to the reference trajectory and is only reliant on measurements that the robot has access to during hardware deployment. 

\textbf{Student Architecture:} To process the observations, the student policy has the form of a convolutional neural network (CNN) to process the depth image, a transformer that fuses the CNN embeddings and proprioception history data, and finally a mixture density network (MDN) that outputs a bi-modal action distribution. The use of the CNN and transformer was inspired by \cite{zhang2026rpl}, while the MDN output was added to 
enable the student policy to disambiguate and choose between different stepping modes (e.g., one foothold instead of another or stepping with the left or right foot). 


\begin{remark}
The function $V$ is a control Lyapunov function only under the conditions stated in Appendix~\ref{apdx:clf-rl} for a fully actuated, flat-footed single-support phase. Our reference-guided rewards are heuristic for the full walking task: the single-support result does not cover impacts, contact changes, or the complete hybrid locomotion cycle. The finite-difference estimate of $\dot V$ also introduces noise into the reward, in addition to the Monte Carlo estimation noise in PPO. Thus, the reward construction does not establish a stability or safety guarantee for the learned policy.
\end{remark}


\section{Training Setup} \label{sec:setup}

Next we provide important information regarding our training setup. Please see the appendix for additional information regarding compute, hyperparameters and domain randomization distributions.

\textbf{Teacher Training:} The teacher policy operates with privileged access to the reference commands, $\mb{y}_\textup{ref}$, which are recomputed at every new step contact. The reference commands and true system states, $\mb{y}_\textup{sim}(t)$, are used to construct $V(\bs{\eta}) $ as in \eqref{eq:clf_rewards} from a pre-computed positive definite matrix $\mb{P}$. The derivative $\frac{dV}{dt}(\bs{\eta}, \mb
{a})$ is approximated using three-point backward difference during training. These are used to evaluate the Lyapunov-guided heuristic rewards \eqref{eq:clf_rewards} which are added to the general RL rewards \eqref{eq:reward}.

The other rewards include holonomic stance-foot rewards penalizing displacement and velocity of the stance foot from its contact location, torque minimization reward, joint limits reward, and action smoothness. The holonomic stance-foot reward is included to incentivize the learned policy to satisfy the underlying assumptions of the HLIP model. The termination conditions are: timeout, fall over (i.e., $>50^\circ$ tilt), pelvis too low, foot height too low, and MPC infeasibility. The domain randomization parameters can be found in Table \ref{tab:domain_randomization} in the Appendix. 

\begin{figure}[t]
    \centering
    \includegraphics[width=0.9\linewidth]{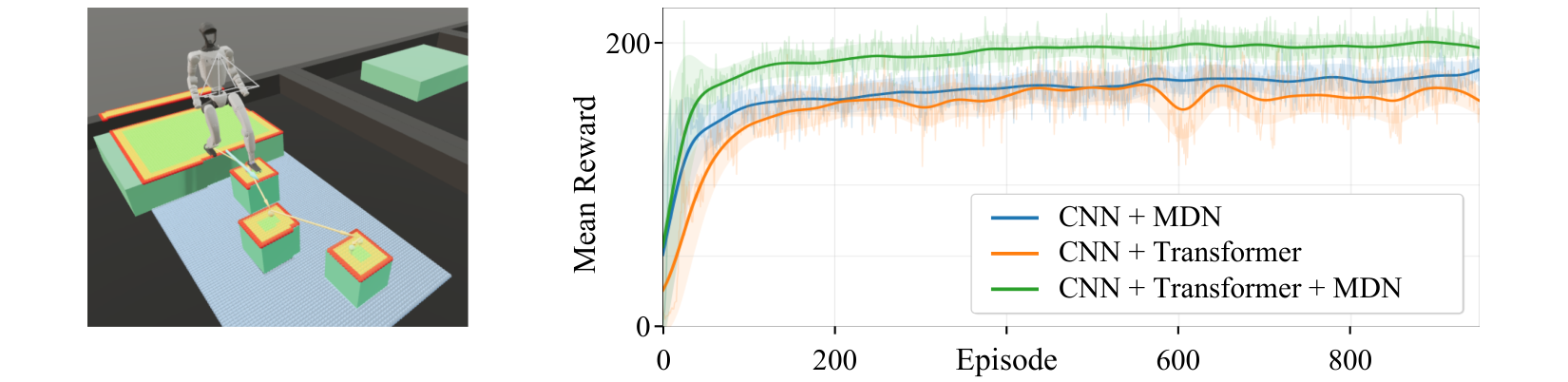}
    \caption{\textbf{Simulation Environment.} \textbf{(Left)} Unitree G1 in $\texttt{mjlab}$: the model-based planner uses the true height map (red/yellow) for a safe foothold plan (yellow arrows); the white camera view near the torso is used for student training. \textbf{(Right)} Fixed teacher $\bs{\pi}_{T,\theta}$, 1000 student-training episodes: ours $\bs{\pi}_{S,\phi}$ \textcolor{olive}{(green)} consistently outperforms transformer \textcolor{blue}{(blue)} and MDN \textcolor{orange}{(orange)} ablations. Training uses distillation loss; student rollouts are evaluated with the full teacher training reward $\mathcal{R}(\mb{s}_k,\mb{a}_k)$.}
    \label{fig:sims}
    \vspace{-1em}
\end{figure}

The teacher is trained on a procedurally generated terrain consisting of two end platforms connected by a sequence of stepping stones, as seen in Fig. \ref{fig:sims}. A curriculum is designed where stone dimensions, gaps, and height variations vary with increasing difficulty. Additionally, random side-by-side stone pairs are also generated to create left/right foot-placement decisions. 

\textbf{Student Distillation:} After training the teacher policy, we distill it into the student policy that no longer has access to privileged reference trajectories or ground-truth height map information and instead only has access to a body-mounted depth camera that is facing forwards and down. The student policy input consists of both this 32$\times$24 pixel depth image and a 5-step history of proprioception data. 
Our student policy is trained using the distillation loss \eqref{eq:distillation_loss} specific to the MDN output. The depth observations used to train the student policy are corrupted with noise, pixel dropout, and a delay to approximate real sensor values.

\section{Experimental Results} \label{sec:result}

We evaluate our model-assisted reinforcement learning framework through simulation validation trials on the Unitree G1 humanoid robot. When compared to standard model-free approaches exemplified by \cite{zhang2026rpl}\footnote{We are unable to perform a direct comparison with \cite{zhang2026rpl} since 
no code has been published. The included results are our best approximation of their method according to the details in the preprint.}, our evaluation demonstrates that model guidance improves both sample efficiency in training and locomotion smoothness at deployment. The distilled student policy succeeded in $66.70\%$ of 1024 trials on a randomized four-stone corridor.

\subsection{Simulations}

\textbf{Ablation Study:} We first perform an ablation study to test the utility of the transformer network and the MDN output. Specifically, we compare our method against two ablations: one with the transformer replaced with an MLP and another with the MDN replaced with an MLP trained to perform regression (so that the output is deterministic). We use a fixed teacher policy, $\bs{\pi}_{T, \theta}$ to train each student. The models with MDN layers are trained using a negative log-likelihood loss as in \eqref{eq:distillation_loss} and the model with the deterministic MLP output is trained with a Huber loss as in \cite{dai2026walk}. The results are presented in Fig. \ref{fig:sims} and show that the transformer and MDN both improve student performance as represented by their ability to generate reward after distillation.

\textbf{Reference-reward ablation:} We compare the Lyapunov-guided heuristic reward with quadratic reference tracking (``Quadratic'') and no reference tracking (``No Ref.''). Across 1024 episodes per policy, our method travels farther and survives longer. Quadratic tracking has lower pelvis motion variance, but terminates after approximately one third of the distance traveled by our method. No Ref. has no privileged foothold information because that information is contained in the reference.
\begin{table}[htbp]
    \centering
    \small
    \setlength{\tabcolsep}{4pt}
    \renewcommand{\arraystretch}{0.90}
    \setlength{\belowcaptionskip}{4pt}
\begin{tabular}{lccccc}
\toprule
 & \multicolumn{3}{c}{Table~\ref{tab:reference_reward_ablation}.A} & \multicolumn{2}{c}{Table 1.B} \\
\cmidrule(lr){2-4}\cmidrule(lr){5-6}
 & Our method & Quadratic & No Ref. & Our method & Single integrator only \\
\midrule
\textbf{(a)} & {\boldmath$3.43\pm0.04$} & $1.02\pm0.01$ & $0.747\pm0.003$ & {\boldmath$3.43\pm0.04$} & $0.933\pm0.013$ \\
\textbf{(b)} & {\boldmath$6.88\pm0.07$} & $2.79\pm0.03$ & $1.50\pm0.01$ & {\boldmath$6.88\pm0.07$} & $1.992\pm0.022$ \\
\textbf{(c)} & {\boldmath$1.40\pm0.07$} & $1.94\pm0.03$ & $6.20\pm0.07$ & {\boldmath$1.40\pm0.07$} & $1.819\pm0.026$ \\
\textbf{(d)} & {\boldmath$0.440\pm0.022$} & $0.460\pm0.010$ & $2.125\pm0.030$ & $0.440\pm0.022$ & {\boldmath$0.409\pm0.011$} \\
\textbf{(e)} & $0.341\pm0.008$ & {\boldmath$0.296\pm0.005$} & $2.914\pm0.026$ & $0.341\pm0.008$ & {\boldmath$0.281\pm0.004$} \\
\textbf{(f)} & $0.173\pm0.005$ & {\boldmath$0.102\pm0.002$} & $1.437\pm0.019$ & $0.173\pm0.005$ & {\boldmath$0.048\pm0.001$} \\
\textbf{(g)} & $0.243\pm0.006$ & {\boldmath$0.182\pm0.009$} & $0.384\pm0.008$ & $0.243\pm0.006$ & {\boldmath$0.060\pm0.004$} \\
\bottomrule
\end{tabular}
\caption{\textbf{A} \textbf{(Left)} Three policies, 1024 episodes/policy; means $\pm$ 1 SEM, all episodes terminated. \textbf{(a,b)} Distance [m] and time [s] before termination; \textbf{(c)} contact-force variance [$10^4\times\mathrm{N}^2$]; \textbf{(d)} mean-squared contact-force rate [$10^8\times\mathrm{N}^2/\mathrm{s}^2$]; \textbf{(e--g)} pelvis roll/pitch/yaw variance [$10^{-2}\times\mathrm{rad}^2$].  \textbf{B (Right)} Single-integrator ablation (ours: single-integrator + HLIP reference). Means $\pm$1 SEM; \textbf{(a)} forward displacement [m], \textbf{(b)} time to episode end [s]; \textbf{(c--g)} Table~\ref{tab:reference_reward_ablation} metrics/units. \textbf{Best in bold.}}
\label{tab:reference_reward_ablation}
\label{tab:single_integrator_ablation}
    \par\smallskip
    \begin{tabular}{lccccc}

        \toprule
        & $N=1$ & $N=2$ & $N=3$ & $N=4$ & $N=8$ \\
        \midrule
        \textbf{(a)} & $3.153\pm0.066$ & $3.256\pm0.071$ & {\boldmath$3.575\pm0.082$} & $3.388\pm0.082$ & $2.698\pm0.056$ \\
        \textbf{(b)} & $7.057\pm0.138$ & $7.200\pm0.134$ & {\boldmath$7.990\pm0.165$} & $7.935\pm0.178$ & $6.611\pm0.109$ \\
        \textbf{(c)} & $1.609\pm0.038$ & $1.506\pm0.027$ & $1.459\pm0.031$ & {\boldmath$1.386\pm0.029$} & $1.489\pm0.027$ \\
        \textbf{(d)} & $0.450\pm0.010$ & $0.429\pm0.008$ & $0.412\pm0.009$ & {\boldmath$0.387\pm0.008$} & $0.409\pm0.007$ \\
        \textbf{(e)} & $0.349\pm0.007$ & $0.353\pm0.007$ & {\boldmath$0.302\pm0.006$} & $0.317\pm0.006$ & $0.352\pm0.006$ \\
        \textbf{(f)} & $0.267\pm0.015$ & $0.192\pm0.007$ & $0.189\pm0.007$ & $0.165\pm0.007$ & {\boldmath$0.160\pm0.006$} \\
        \textbf{(g)} & $0.435\pm0.018$ & $0.359\pm0.019$ & {\boldmath$0.324\pm0.010$} & $0.328\pm0.012$ & $0.427\pm0.019$ \\
        \bottomrule
    
    \end{tabular}
\caption{Planning-horizon ablation. Means $\pm$ 1 SEM, rounded to three decimals; metrics and units as in Table~\ref{tab:reference_reward_ablation}. \textbf{Best mean in bold.}}
\label{tab:planning_horizon_ablation}
    \vspace{-8pt}
\end{table}

\textbf{MDN multi-modality:} In simulation, we swept the initial CoM lateral offset across 321 values spanning $\pm0.3$\,m and recorded foot touchdown locations. The policy deterministically selects the highest-probability component of its two-component MDN. Fig.~\ref{fig:mdn_modes} shows the resulting spatially distinct touchdown modes.
\definecolor{compA}{HTML}{0072B2}
\definecolor{compB}{HTML}{E69F00}
\begin{figure}[htbp]
    \centering
    \includegraphics[width=0.85\linewidth]{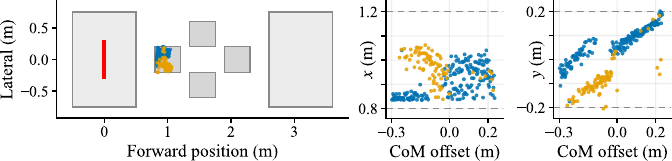}
    \caption{\textbf{(Left)} Stepping-stone/initial CoM conditions \textcolor{red}{(red)}; \textcolor{compA}{blue}/\textcolor{compB}{yellow} touchdowns identify the dominant mixture component (argmax for a median 75\% of each step). Conditioned on successful touchdown, components dominate 69\%/31\% of the runs, respectively. \textbf{(Center/Right)} Forward/lateral touchdown location versus CoM offset for each mode.}
    \label{fig:mdn_modes}
    \vspace{-10pt}
\end{figure}

\begin{figure}[htbp]
    \centering
    \includegraphics[width=\linewidth]{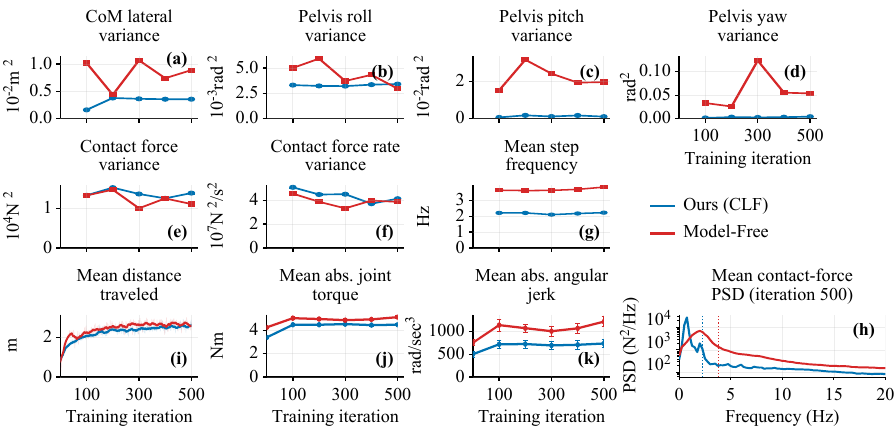}
    \caption{Smoothness comparisons: our model-informed method \textcolor{blue}{(blue)} reduces or matches variances and has contact-force spectral power more concentrated at lower frequencies than the model-free baseline \textcolor{red}{(red)}; shown to 20\,Hz. \textbf{(a)--(h)}: 64 episodes/checkpoint, logged at 50\,Hz; \textbf{(a)--(g)}: $\pm1$ SEM error bars; \textbf{(h)}: dotted mean. Other $x$-axes show distillation training iterations. Teachers: 10,000 model-informed/20,000 model-free episodes. \textbf{(i)--(k)}: 2048 rollouts/iteration; \textbf{(i)}: 20-episode moving average, slightly faster model-free convergence and similar final mean distances.}
    \label{fig:smoothness_rebuttal}
    \label{fig:rpl_comparison}
    \vspace{-1em}
\end{figure}

\textbf{Baseline Comparison:} We compare our method to a model-free PPO baseline similar to \cite{zhang2026rpl}, sharing our teacher-student distillation framework and privileged environmental information but omitting Lyapunov-guided heuristic rewards.  
Both reach similar mean distances over stepping stones, but our teacher converges in 10,000 episodes versus 20,000 for the baseline. The method in \cite{zhang2026rpl} also used a more structured curriculum from flat ground to stepping stones. On one RTX 5090, teacher training and student distillation each took 12 GPU hours (24 GPU hours total) for both our method and the model-free comparison. Reference-trajectory computation raises per-episode cost, making total wall-clock training time comparable despite improved sample efficiency; caching references or reducing height-map density could reduce this primary overhead. Across 2048 rollouts of distilled student policies, our method achieves at least 12\% lower mean absolute joint torque and 39\% lower mean absolute angular jerk after 500 distillation iterations (Fig.~\ref{fig:rpl_comparison}), yielding smoother locomotion (video: \href{https://www.youtube.com/watch?v=pXjy-zPCRPA}{youtube.com/watch?v=pXjy-zPCRPA}).

\subsection{Hardware Experiments}

While our primary results are in simulation, we additionally provide a single proof-of-concept hardware demonstration; we do not claim a success rate for these hardware experiments. On hardware, we iterated deployment until achieving a successful crossing and did not run a fixed-$N$ success-rate protocol. The experimental setup consists of a starting platform, an ending platform, and a sequence of 4 stepping stones arranged in a pseudo-random, alternating left-and-right configuration. This layout forces the robot to tightly control both its forward and lateral foot placements. The stepping stones are all the same height and covered with a rigid wood board to replicate the flat contact surfaces used in simulation. 
The distilled student policy runs entirely on the Unitree G1's Jetson Orin Nx onboard computer and utilizes a chest-mounted Intel Realsense D435i for depth image measurements. The full vision-proprioception pipeline, including the CNN feature extractor and transformer, executes at approximately 50Hz, matching the control loop frequency in simulation.

\section{Limitations}\label{sec:limitations}

Firstly, our approach is limited by the sensory modes of the system. The camera is pointed forwards and down, preventing the ability to learn safe backwards or sideways locomotion. The angle of the camera also limits the ability of the robot to achieve highly dynamic, long-horizon maneuvers that might require planning motions over several ($>$4) steps. Future work will involve actively directing the camera to enable these behaviors. Additionally, the HLIP model does not include a flight phase which means that our reference trajectories could not include jumping. Future work will develop and incorporate alternative reduced-order models to enable this behavior.
Furthermore, while the proposed method improved sampling efficiency, the overall training time was similar. Future work will involve improving per-sample computational efficiency.
Finally, while the policy achieved a successful demonstration on hardware, the proof-of-concept physical deployment was significantly less repeatable than simulation; we attribute this to camera latency and accuracy and seek to achieve improved robustness in the future.


\section{Conclusion} \label{sec:conclusion}

We propose a model-guided reinforcement learning framework to enable vision-based sparse foothold navigation with lateral constraints for a bipedal robot. Our three-step method involves first using simplified dynamics models (discrete-time single integrator and HLIP) to plan a reference trajectory, then using this reference trajectory to guide the training of a privileged teacher control policy, and finally distilling that teacher policy into a vision-based student.
We deploy this method on hardware and show that this model-guided learning framework demonstrates improved sample efficiency and yields smoother locomotion compared to model-free alternatives.


\clearpage
\bibliography{cosner}  

\clearpage

\section{Appendix}
\label{sec:acknowledgments}

\subsection{Artificial Intelligence Acknowledgement}
The authors acknowledge the use of artificial intelligence (AI) technologies in the preparation of this paper. Specifically, large language models were utilized for the following purposes:

\begin{itemize}
    \item \textbf{Code Generation:} AI tools assisted in writing the software, including the simulation setup, training scripts, plot generators, and experimental evaluation.
    \item \textbf{Writing Assistance:} AI was used as a supportive tool to improve the clarity, grammar, and phrasing of the text during the writing and editing process. 
\end{itemize}

All AI-generated outputs were thoroughly reviewed, verified, and edited by the authors, who take full responsibility for the contents of this paper.

\subsection{Experimental Parameters and Reproducibility}
To ensure reproducibility, this section provides comprehensive details regarding the simulation parameters, RL hyperparameters, and hardware configurations utilized across our pipeline. Specifically, Table \ref{tab:training_efficiency} summarizes the simulation framework, the scale of our parallelized environment setup, and the specific hardware infrastructure used to execute the training. Table \ref{tab:domain_randomization} outlines the complete suite of domain randomization ranges applied during the training process. These perturbations include rigid-body variations, sensor noise models, and push forces designed to bridge the sim-to-real gap. These are the same as in \cite{zhang2026rpl}. Finally, Table \ref{tab:training_hyperparameters} provides the exact training hyperparameters used for both components of our learning pipeline: the reinforcement learning teacher policy (trained via PPO) and the subsequent vision-based student policy (distilled from the teacher policy).

\begin{table}[htbp]
    \centering
    \small
    \renewcommand{\arraystretch}{1.25}
    \begin{tabular}{p{0.35\linewidth}p{0.57\linewidth}}
        \toprule
        \textbf{Configuration / Metric} & \textbf{Details \& Infrastructure Specifications} \\ 
        \midrule
        \textbf{Simulation Environment} & \texttt{mjlab} framework \cite{zakka2026mjlablightweightframeworkgpuaccelerated} via MuJoCo simulation environment \cite{todorov2012mujoco} \\ 
        \textbf{Learning Library} & GPU-accelerated \texttt{rsl-rl} library \cite{schwarke2025rslrllearninglibraryrobotics} \\ 
        \textbf{Parallelization} & 4,096 parallelized simulated environments \\ 
        \textbf{Hardware Infrastructure} & Single NVIDIA RTX 5090 GPU \\ 
        \textbf{Pipeline Training Time} & 24 GPU hours on an RTX 5090: 12 hours for teacher training and 12 hours for student distillation \\ 
        \bottomrule
    \end{tabular}
    \caption{Simulation environment, training efficiency, and computational infrastructure details.}
    \label{tab:training_efficiency}
\end{table}

\begin{table}[htbp]
    \centering
    \renewcommand{\arraystretch}{1.15} 
    \begin{tabular}{ll}
        \toprule
        \textbf{Parameter} & \textbf{Value} \\
        \midrule
        Friction Coefficients & $\mathcal{U}(0.5, 1.25)$ \\ 
        Link Mass & $\mathcal{U}(0.9, 1.2) \times$ default mass \\
        Base Additional Mass & $\mathcal{U}(-1.0, 3.0)$ kg \\
        Encoder Bias  & $\mathcal{U}(-0.015, 0.015)$ rad \\
        Base CoM Offset & $\Delta x \sim \mathcal{U}(-0.025, 0.025)$ m \\
                        & $\Delta y \sim \mathcal{U}(-0.05, 0.05)$ m \\
                        & $\Delta z \sim \mathcal{U}(-0.05, 0.05)$ m \\
        Motor Effort Limits & $\mathcal{U}(0.85, 1.1) \times$ default \\
        Joint Damping & $\mathcal{U}(0.75, 1.25)\times$ default \\
        Joint Friction & $\mathcal{U}(0.0, 0.04)$ \\
        Push Perturbation: & $x \sim \mathcal{U}(-1, 1) \frac{\textup{m}}{\textup{s}}$ \\
        \quad random velocity & $y\sim \mathcal{U}(-1, 1) \frac{\textup{m}}{\textup{s}}$ \\ 
        \quad added every 10-15s & $(\phi, \theta, \psi)\sim \mathcal{U}^3(-0.4, 0.4) \frac{\textup{rad}}{\textup{sec}} $ \\
        Control Delay & $\mathcal{U}(0, 20)$ ms \\
        Depth Camera Translation & $x \sim \mathcal{U}(-0.025, 0.025)$ m \\
                                & $y \sim \mathcal{U}(-0.025, 0.025)$ m \\
                                & $z \sim \mathcal{U}(-0.025, 0.025)$ m \\
        Depth Camera Rotation (Euler) & $\phi \sim \mathcal{U}(-2.5, 2.5)^\circ$ \\
                                      & $\theta \sim\mathcal{U}(-3.0, 3.0)^\circ$ \\
                                      & $ \psi \sim \mathcal{U}(-2.5, 2.5)^\circ$ \\
        Depth Camera FOV & $\Delta$FOV$\sim \mathcal{U}(-2.0, 2.0)^\circ$ \\
        Pixel Dropout & $5\%$ \\
        Depth Noise & $\sigma_d = 0.1 \cdot $ depth \\
        \bottomrule
    \end{tabular}
    \caption{Domain Randomization. The values were chosen to align with the values in \cite{zhang2026rpl}.}
    \label{tab:domain_randomization}
\end{table}

\begin{table}[htbp]
    \centering
    \small
    \renewcommand{\arraystretch}{1.15}
    \begin{tabular}{p{0.46\linewidth}p{0.46\linewidth}}
        \toprule
        \textbf{Lyapunov-guided teacher PPO} & \textbf{MDN Distillation} \\
        \midrule
        \textbf{Network:} MLP actor-critic, $(512, 256, 128)$ hidden units, ELU activation \newline
        \textbf{Observation normalization:} enabled \newline
        \textbf{Actor initial std:} $1.0$ \newline
        \textbf{Learning rate:} $5\times 10^{-5}$ \newline
        \textbf{LR schedule:} adaptive \newline
        \textbf{PPO epochs:} $5$ \newline
        \textbf{Mini-batches:} $4$ \newline
        \textbf{Clip parameter:} $0.2$ \newline
        \textbf{Value loss coefficient:} $1.0$ \newline
        \textbf{Entropy coefficient:} $0.008$ \newline
        \textbf{Discount factor:} $\gamma = 0.99$ \newline
        \textbf{GAE parameter:} $\lambda = 0.95$ \newline
        \textbf{Target KL:} $0.01$ \newline
        \textbf{Max gradient norm:} $1.0$ \newline
        \textbf{Rollout length:} $24$ steps/env \newline
        \textbf{Training iterations:} $10000$
        &
        \textbf{Student network:} CNN--Transformer--MDN, ELU activation \newline
        \textbf{Teacher network:} Lyapunov-guided PPO policy \newline
        \textbf{Observation normalization:} enabled \newline
        \textbf{MDN modes:} $2$ \newline
        \textbf{Minimum std:} $10^{-3}$ \newline
        \textbf{Log-std range:} $[-3, 2]$ \newline
        \textbf{Inference mode:} \texttt{top\_mode\_mean} \newline
        \textbf{Optimizer:} Adam \newline
        \textbf{Learning rate:} $5\times 10^{-4}$ \newline
        \textbf{Distillation epochs:} $10$ \newline
        \textbf{Gradient length:} $2$ \newline
        \textbf{Teacher samples:} $8$ \newline
        \textbf{Teacher std scale:} $0.1$ \newline
        \textbf{Max gradient norm:} $2.0$ \newline
        \textbf{Rollout length:} $120$ steps/env \newline
        \textbf{Training iterations:} $1500$ \\
        \bottomrule
    \end{tabular}
    \caption{Training hyperparameters for the Lyapunov-guided PPO teacher and the MDN distillation student.}
    \label{tab:training_hyperparameters}
\end{table}

\subsection{Conditions for a Control Lyapunov Function} \label{apdx:clf-rl} 

Consider the dynamics of the single-support phase for a bipedal robot as given in \cite{li2026clfrl}: 
\begin{align}
    \mb{D}(\mb{q}) \Ddot{q} + \mb{H}(\mb{q}, \dot{\mb{q}}) = \mb{B}\mb{u} + \mb{J}_h(\mb{q})^\top \mb{F} \label{eq:fully-actuated-humanoid}\\
    \mb{J}_h(\mb{q})\Ddot{q} + \dot{\mb{J}}_h (\mb{q}, \dot{\mb{q}}) \dot{\mb{q}} = 0 \label{eq:constraint-for-humanoid}
\end{align}
where $\mb{D}(\mb{q}): \mathcal{Q}\to \R^{n_q\times n_q} $ is the mass-intertia matrix, $\mb{H}: \mathcal{Q}\times T\mathcal{Q} \to \R^{n_q}$ contains the Coriolis and gravity terms, and $\mb{B} \in \R^{n_q, n_u}$ is the actuation matrix. The Jacobian of the holonomic contact constraint is $\mb{J}_h(q) \in \R^{h\times n_q}$ and the associated constraint wrench is $\mb{F} \in \R^h$

\begin{proposition}[{\cite[Prop. 1]{li2026clfrl}}]
    Consider a fully actuated humanoid in flat-footed single support whose dynamics satisfy \eqref{eq:fully-actuated-humanoid}, the holonomic contact constraint \eqref{eq:constraint-for-humanoid}, and friction-cone constraints. If the CARE uses $\mb{Q}\succ 0$ and the virtual-constraint Jacobian $\partial y/\partial q$ is invertible, then $V(\bs{\eta})=\bs{\eta}^\top\mb{P}\bs{\eta}$, with $\mb{P}\succ0$ given by the CARE for the feedback-linearized error dynamics, is a control Lyapunov function for the full nonlinear dynamics during single support.
\end{proposition}
This conditional result applies to the continuous, flat-footed single-support phase. It does not establish Lyapunov decrease across impacts or other contact changes, nor a guarantee for the complete hybrid walking motion used here.

\end{document}